\documentclass[sigconf]{acmart}
\usepackage{pifont, alltt, multirow, colortbl, graphicx, subcaption, caption, enumitem, setspace}
\usepackage[tableposition=top]{caption}

\AtBeginDocument{%
  \providecommand\BibTeX{{%
    \normalfont B\kern-0.5em{\scshape i\kern-0.25em b}\kern-0.8em\TeX}}}

\setcopyright{acmcopyright}
\copyrightyear{2022}
\acmYear{2022}
\acmDOI{XXXXXXX.XXXXXXX}

\acmConference[MET '22]{7th International Workshop on Metamorphic Testing}{May 21--29,
  2022}{Pittsburgh, PA}

\acmPrice{15.00}
\acmISBN{978-1-4503-XXXX-X/18/06}

\begin{document}

\title{Fairness Evaluation in Deepfake Detection Models using Metamorphic Testing}

\author{Muxin Pu}
\affiliation{%
  \institution{Monash University Malaysia}
  \state{Selangor}
  \country{Malaysia}}
\email{mpuu0001@student.monash.edu}

\author{Meng Yi Kuan}
\affiliation{%
  \institution{Monash University Malaysia}
  \state{Selangor}
  \country{Malaysia}}
\email{mkua0001@student.monash.edu}

\author{Nyee Thoang Lim}
\affiliation{%
  \institution{Monash University Malaysia}
  \state{Selangor}
  \country{Malaysia}}
\email{nlim0019@student.monash.edu}

\author{Chun Yong Chong}
\affiliation{%
  \institution{Monash University Malaysia}
  \state{Selangor}
  \country{Malaysia}}
\email{chong.chunyong@monash.edu}

\author{Mei Kuan Lim}
\affiliation{%
  \institution{Monash University Malaysia}
  \state{Selangor}
  \country{Malaysia}}
\email{lim.meikuan@monash.edu}

\renewcommand{\shortauthors}{Pu et al.}

\begin{abstract}


Fairness of deepfake detectors in the presence of anomalies are not well investigated, especially if those anomalies are more prominent in either male or female subjects. The primary motivation for this work is to evaluate how deepfake detection model behaves under such anomalies. However, due to the black-box nature of deep learning (DL) and artificial intelligence (AI) systems, it is hard to predict the performance of a model when the input data is modified. Crucially, if this defect is not addressed properly, it will adversely affect the fairness of the model and result in discrimination of certain sub-population unintentionally. Therefore, the objective of this work is to adopt metamorphic testing to examine the reliability of the selected deepfake detection model, and how the transformation of input variation places influence on the output. We have chosen MesoInception-4, a state-of-the-art deepfake detection model, as the target model and makeup as the anomalies. Makeups are applied through utilizing the Dlib library to obtain the 68 facial landmarks prior to filling in the RGB values. Metamorphic relations are derived based on the notion that realistic perturbations of the input images, such as makeup, involving eyeliners, eyeshadows, blushes, and lipsticks (which are common cosmetic appearance) applied to male and female images, should not alter the output of the model by a huge margin. Furthermore, we narrow down the scope to focus on revealing potential gender biases in DL and AI systems. Specifically, we are interested to examine whether MesoInception-4 model produces unfair decisions, which should be considered as a consequence of robustness issues. The findings from our work have the potential to pave the way for new research directions in the quality assurance and fairness in DL and AI systems.
\end{abstract}

\ccsdesc[500]{Software and its engineering~Software testing and debugging}

\keywords{Metamorphic testing, fairness testing, robustness testing, oracle problem}

\maketitle

\section{Introduction}
With the advancement of sophisticated DL and AI systems/models coupled with high computational powers of hardware, various application of AI models such as deepfake has started to emerge and gain popularity. While most researchers focus mainly on improving the accuracy of their proposed models, there is a growing need to investigate the robustness and fairness of AI and DL models, which can be viewed as the most essential non-functional requirements in any well-designed AI and DL models \cite{mehrabi2021survey}. 
Deepfakes are created using techniques that superimpose a target person's face onto a source person in a video \cite{9544522}. The issues derived from deepfake are becoming more serious nowadays, as deepfake generation methods could potentially allow its user to fabricate lies and spread misinformation for malicious purposes. Deepfake detection techniques are still in their infancy and a variety of approaches have been proposed and tested, albeit on fragmented datasets. 
While there is a considerable interest and rapid advancement of deepfake detection models, it is not accompanied with reliable and large enough datasets to evaluate the performance of new and innovative deepfake detection models (modified benchmark dataset of deepfake images and videos) \cite{9544522}.

On the other hand, new deepfake detection techniques mostly rely on flaws in deepfake generation pipelines. However, in adversarial contexts, where attackers often try not to disclose the working principles of their attacks, the way how adversarial examples/datasets are generated is not always available. To simplify, the forms of adversarial contexts are not clearly structured underneath the complex layers of both deepfake generation and detection models, and hence, it is not always true that we can verify the existence of a certain oracle problem by the traditional testing strategies. The challenge of deepfake detection has become more complex as a result of recent research on adversarial perturbation attacks to trick DNN-based detectors \cite{9544522}.


In most existing research deepfake detection models, they are trained on well-chosen or clear frontal image of the subject without any foreign artifacts. How would the model perform in situations where the training images are not in ideal conditions (e.g. consisting of foreign artifacts, more than one face, makeup and accessories on the subject face, artificial noises, etc.) are not well studied due to the lack of labelled ground truth.

Metamorphic testing is one of the widely used approaches to test and evaluate the robustness of AI and machine learning algorithms, specifically to deal with the oracle problem. In this work, we aim to conduct metamorphic testing to aid the evaluation of robustness and fairness of MesoInception-4 model against adversarial attacks by adding makeup to the testing images. We would like to emphasise that the application of makeup should not be treated as deepfakes as it only modifies the RGB values of the face images and does not transfer the facial features of one person to another. Often, in the real world, makeups are very commonly applied by both genders, and hence, when detecting whether the input image is a deepfake or not, the application of makeup should not be considered as unexpected or borderline cases. The application of makeup on the testing images could be treated as \textit{"adversarial attack"}, which can potentially reveal some hidden bugs, robustness, or fairness issues in the examined deepfake detection model. 

There is therefore a need to investigate the performance of the model on input images that contains foreign artifacts. On the other hand, we are interested to evaluate the fairness of deepfake detection models by observing if there are discriminated decision caused by biased learning models, especially in the case where deepfake detectors are performing better on one gender over the other. This is because when a model is constructed to make decisions with high accuracy, \textit{fairness} would be an essential non-functional requirement of the model. The term \textit{fairness} in this study means that the selected model should not provide unfair decisions based on a specific gender. 

The goal of this paper is to utilise metamorphic testing to aid in the evaluation of robustness and fairness of deepfake detection model. The following are the specific goals of this paper: 
\begin{itemize}[topsep=0pt]
    \item To propose metamorphic relations that are capable to examine the robustness of MesoInception-4 model against adversarial attack.
    \item To examine the potential existence of gender bias in \\ MesoInception-4 model.
\end{itemize}


\section{Background}
\subsection{Deepfake Detection}


The success and advancement of deepfake is mainly brought by the the emergence of convolutional neural networks (CNN) and generative adversarial networks (GANs), which are the cores of deepfake generation techniques. By having more precise facial landmark detection, segmentation, and pose estimation algorithms, deepfake technologies make it harder for humans and computers to detect forgeries. The impact of wrongfully used deepfake technology can be disastrous. Hence, researchers across the globe are working in hands to propose various countermeasures against it from time to time. Earlier attempts of deepfake detection focus on identifying the inconsistencies of the fake video synthesis process and handcrafted artifacts obtained from the images \cite{9544522}. Methods that identify inconsistencies are usually useful for detecting the conventional deepfake conducted by \textit{blending method} which fuses the source face into the target facial region of interest (ROI) built from the facial landmarks collection \cite{ding_raziei_larson_olinick_krueger_hahsler_2020}. 

Even though many deepfake detection models have been proposed over the years with promising results, there are limited research conducted to evaluate the solutions' performance from the gender perspective. The popular deepfake datasets such as FaceForensics++, Korean Deepfake Detection (KoDF), Deepfake Detection Challenge (DFDC) and Celeb-DF datasets which are commonly used consist of data source from a combination of both genders regardless of their cardinality. Since AI technologies typically learn from human-labelled datasets during the training phase, any potential biases from the human labeller could be imposed on the decisions made by the AI model unintentionally. As such, deepfake detection models are also subjected to the potential issue of possessing gender bias.

\subsection{Fairness of Deepfake Detector}
 Bias is defined as \textit{‘the inclination or prejudice of a decision made by an AI system which is for or against one person or group, especially in a way considered to be unfair’} \cite{https://doi.org/10.1002/widm.1356}. Several factors may cause biases in AI models, including socio-technical factors such as data generation, data collection, and institutional bias. Computer vision models in particular, are also susceptible to harmful and pervasive prejudice. The work conducted by Weiss et. al. has shown that racial bias exists in deepfake detectors when the three deepfake detector systems tested do not perform equally on a general population with a noticeable error rate of 10.7 percent \cite{trinh2021examination}. Many practitioners and researchers have been putting in efforts to mitigate biases in AI, typically through innovative ways to alter the AI model's training process, such as removing input dataset bias during the preprocessing phase \cite{https://doi.org/10.1002/widm.1356}. Another method to mitigate biases is the introduction of weights into the model purposely to explicitly incorporate the model's discrimination behaviour \cite{https://doi.org/10.1002/widm.1356}. Additionally, the model’s internals or prediction can also be modified directly through the white-box or black-box approach if necessary to mitigate biases \cite{https://doi.org/10.1002/widm.1356}. Despite all the efforts conducted, mitigating biases in deepfake detection model is still a challenge task due to the insufficient concrete proof, especially the access to oracle during the evaluation of deepfake detectors. In fact, the complex underlying network of the unsupervised DL models cause the retrieval of the oracle information in the form of dataset \textit{'ground truth'} remains an ongoing challenge.

\subsection{Metamorphic Testing}
Due to the fact that deep learning-based AI technologies do not come with explicit input and output specifications that can be simply tested and verified, it is therefore crucial to evaluate the performance of DL models from various perspectives to ensure they are generalised, stable, and consistent \cite{StackedGenerativeAdversarialNetworksEvaluation}. Metamorphic testing (MT) is therefore suited to be used to assess the potential vulnerability of deepfake detector, which then in turn offers a research direction to create a more robust and reliable deepfake detector. Previous works have successful applied metamorphic testing to evaluate the robustness of face recognition models \cite{MTonMultiLevelFaceRecognitionModel}. In this work, we aim to evaluate the robustness of deepfake detection models by perturbating the input images to increase their feminine features/characteristics and observe if the detection models can still retain their original evaluation after the perturbation. In return, we will be able to shed some light regarding the potential gender bias of deepfake detection models (if any), and offer some suggestions to minimise such biasness. 




\section{Related Work}
The work by Trinh et. al. \cite{trinh2021examination} performed experiments to investigate the performance of three popular deepfake detection models on the widely used Face Forensics++ dataset and Blended Images (BI) dataset. The goal was to identify whether DL models have systematic discrimination towards certain racial subgroups. The three deepfake detection models are MesoInception-4 \cite{MesoNet}, Xception \cite{rossler2019faceforensics}, and Face X-Ray \cite{li2020face} while the subgroups involve all possible combinations of male and female groups with Caucasian, African, Asian, and Indian racial groups. Data balancing, cross evaluation, and audit were carried out constantly to cross-test the models’ generalisability across datasets with unknown manipulations and to increase the findings’ reliability. From the experimental results, a large disparity in predictive performance across races were found. In general, the authors found out that female Asian or female African are more likely to be mistakenly labelled as fake compared to male Caucasian and females have a higher probability of being mistakenly identified as fake compared to males. Model performance wise, Face X-Ray with BI performs best on Caucasian faces, especially on male Caucasian while Mesonet and Xception perform best on Indian faces \cite{trinh2021examination}. These findings demonstrate how different deepfake detectors discriminate across gender and race by showing gaps in error rates and accuracy across different intersectional groups.

From the model testing perspective, MT has become an effective approach to test the robustness of AI systems. The work by  Wang and Su introduced metaOD, an MT system for object detectors (OD) \cite{ODMT}. Object instances were inserted naturally into background images and metamorphic conditions are designed by comparing the OD results between the original and synthetic images. The experiment successfully revealed tens of thousands of detection failures \cite{ODMT}. Besides, Mekala et. al. built black-box multi-step Projected Gradient Descent (PGD) attacks against Google’s facial recognition system, FaceNet \cite{MTonMultiLevelFaceRecognitionModel}. They propose a metamorphic defense pipeline using combinations of non-linear image transformations such as JPEG compression and Poisson noise perturbations. The best adversarial detection accuracy found is 96\%. Moving on, experiments conducted by Neekhara et al., found the salient regions of the popular Deepfake Detection Challenge (DFDC) dataset on different deepfake detectors are the edges. Adversarial attacks on those regions are discovered to be transferable across different deepfake detectors \cite{AdvAttackonDeepfakeDetector}. The authors then created more accessible attacks relying on Universal Adversarial Perturbations to evaluate against the winning entries of DFDC dataset \cite{AdvAttackonDeepfakeDetector}. These common attacks were proven to be able to bypass multiple deepfake detectors including EfficientNet B7 \cite{EfficientNet} and XceptionNet \cite{XceptionNet}.  

Based on our knowledge, there is no targeted research that performs MT on deepfake detection models to identify potential gender bias. Hence, motivated by the results from Trinh et. al., our work primarily focuses on utilizing MT as a way to evaluate MesoNet, specifically its variation, MesoInception-4, on its gender fairness. The model is particularly chosen due to its efficient and fast binary compact facial forgery detection characteristic with a stable and consistent performance \cite{szegedy2014going}. With an accuracy of up to 96\% reported, this guarantees a faster testing cycle while ensuring a rather reliable result, as shown in Table \ref{tab:score}. On the other hand, MT is chosen as the software testing technique as it has the fault detection capability of effectively alleviating the test oracle problem, commonly faced when testing AI models.

\begin{table}[t]
    \centering
    \caption{Classification scores of MesoNets on Deepfake and Face2Face (compressed at rate 23) datasets, using image aggregation.}
    \label{tab:score}   
    \begin{tabular}{lll}
    \hline
    \multicolumn{3}{l}{Classification Score on Different Datasets (\%)}                       \\ \hline
    \multicolumn{1}{l|}{Network}         & \multicolumn{1}{l|}{Deepfake} & Face2Face (23) \\ \hline
    \multicolumn{1}{l|}{Meso-4}& \multicolumn{1}{l|}{96.9}& 95.3\\ 
    \multicolumn{1}{c|}{MesoInception-4} & \multicolumn{1}{l|}{98.4} & 95.3 \\ \hline
    \end{tabular}
\end{table}

\section{Methods}
To build the ground truth, we trained the MesoInception-4 model following the specifications given by the original authors \cite{MesoNet} using the Adam optimiser with a batch size of 75, and an initial learning rate of $10^3$, where the value is divided by 10 in every 1000 iterations till it reaches $10^6$. Random transformations such as zoom, horizontal flips, and rotation are applied to ensure the generalisation of the model.

In order to obtain a reliable benchmark, we start our evaluation by testing the MesoInception-4 model using our rebalanced images dataset, which only contains images without any transformations. Throughout this study, metamorphic relations are incrementally and iteratively derived based on our intentions at different stages, similar to the work by Park et. al. \cite{StackedGenerativeAdversarialNetworksEvaluation}. Test cases derived from the proposed metamorphic relations are created by adding makeup (metamorphic transformation/perturbations) to the testing images such as eyeshadows, eyeliners, lipstick, and blushes. This is because makeups are deemed common and occur often in the real-world and we are interested to observe a stable and robust performance of the MesoInception-4 model in the presences of anomalies. 

\begin{figure}[b]
    \centering
    \includegraphics[width=0.47\textwidth]{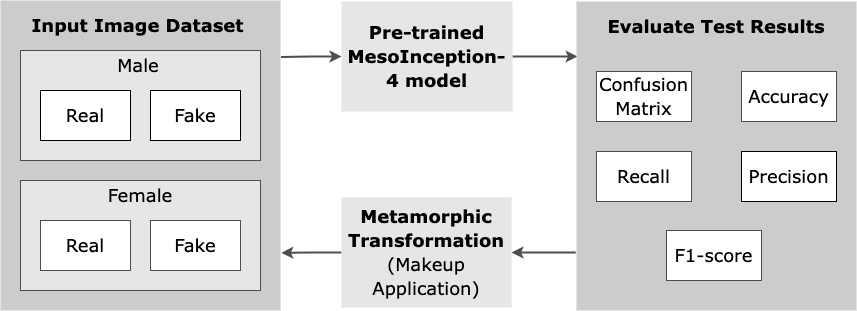}
    \caption{Metamorphic testing workflow used in this study.}
    \label{fig:methodFlow}
\end{figure}

Next, we attempt to verify if the metamorphic relations can still hold with the introduction of perturbed images. To gauge the fairness of the model, we split the testing dataset based on gender and evaluate the model, since it is necessary to analyse the test results by applying unbiased comparisons to examine the pattern of variation. As a result, we can evaluate the model's performance in a smooth controllable manner. Figure \ref{fig:methodFlow} illustrate the overall methodologies used for this study.

To verify the reliability of our experiments, different evaluation metrics such as confusion matrix, accuracy, precision, recall, and F1-score are computed to analyse the test results scientifically.

\section{Experiments}
\subsection{Datasets}
The dataset we used for this study is FaceForensics \cite{rossler2019faceforensics}, which contains the original videos associated with their forged videos generated by Face2Face technique, and all videos with a compression rate of 23 (light). We transform all videos in the dataset to images by closely aligning with the faces using the pipeline provided by Afchar et. al. \cite{MesoNet}. Each extracted image is strictly treated as a single-face image, but there are various postures involved in these images. 

To avoid gender bias triggered by the biased dataset, the dataset is classified and rebalanced based on gender before it is split into the training, validation, and testing sets with the train-test split ratio of 3:2, where 10\% of the training set is being used as the validation set. As a result, we have a total of 3168 sample images for training, 1930 for testing, and 352 for validation. Note that the number of images are divided equally for males and females at each stage. We have also removed some dataset where makeup cannot be applied to the target objects, such as images with non-frontal faces or when there are no faces being detected. 



\subsection{Evaluation Metrics}
Applying the same approach of quantifying the detection performance and obtaining an unbiased comparison is necessary for us, since MesoInception-4 model is designed to have two output categories: 1.) images classified as deepfake or 2.) images classified as non-deepfake (real images). Most error measures only tell us the total error, but to determine individual instances of errors in the model is important in our work. As such, the confusion matrix is useful to show a table layout of the different set of predictions, helping us to visualise its outcomes and offers us a tool to calculate the accuracy, precision, and F1-score for each experiment. 

\begin{figure}[h]
    \centering
    \begin{tabular}{clcc}
    \multicolumn{2}{l}{} & \multicolumn{2}{c}{Prediction} \\ \cline{3-4}
    \multicolumn{1}{l}{} & &
    \multicolumn{1}{l}{Fake} & 
    \multicolumn{1}{l}{Real} \\
    \multicolumn{1}{c|}{} &                      & \multicolumn{1}{c|}{\cellcolor[HTML]{9698ED}{\color[HTML]{FFFFFF} }} & \cellcolor[HTML]{FFCE93} \\
    \multicolumn{1}{c|}{} &
      \multirow{-2}{*}{Fake} &
      \multicolumn{1}{c|}{\multirow{-2}{*}{\cellcolor[HTML]{9698ED}{\color[HTML]{FFFFFF} TP}}} &
      \multirow{-2}{*}{\cellcolor[HTML]{FFCE93}FN} \\ \cline{3-4} 
    \multicolumn{1}{c|}{} & \multicolumn{1}{c}{} & \multicolumn{1}{c|}{\cellcolor[HTML]{6665CD}{\color[HTML]{FFFFFF} }} & \cellcolor[HTML]{FFCC67} \\
    \multicolumn{1}{c|}{\multirow{-4}{*}{Actual}} &
      \multicolumn{1}{c}{\multirow{-2}{*}{Real}} &
      \multicolumn{1}{c|}{\multirow{-2}{*}{\cellcolor[HTML]{6665CD}{\color[HTML]{FFFFFF} FP}}} &
      \multirow{-2}{*}{\cellcolor[HTML]{FFCC67}TN}
    \end{tabular}
    \caption{The confusion matrix used for this study.}
    \label{fig:confuix_metric}
\end{figure}

\noindent MesoInception-4 model provide 4 outcomes shown in \textbf{Figure \ref{fig:confuix_metric}}: 
\begin{itemize}[topsep=0pt]
    \item True positive (TP): the number of correct deepfake images that are correctly identified as deepfake images.
    \item False positive (FP): the number of non-deepfake images incorrectly identified as deepfake images.
    \item True negative (TN): the number of non-deepfake images correctly identified as real images.
    \item False negative (FN): the number of deepfake images incorrectly identified as real images.
\end{itemize}

Accuracy is obtained by calculating the number of all correct predictions divided by the total number of test samples. It has an upper limit of 1 and a lower limit of 0, and is an indispensable metrics to evaluate the performance of a DL model. 
\begin{center}
    \( Accuracy (\%) = \frac{TP + TN }{ TP + FP + FN + TN} \times 100 \) 
\end{center}
\noindent Precision and recall are also important for evaluating the performance of a model. Precision tells us out of all the negative predicted, what percentage is truly positive. Finally, recall is used to show the percentage of predicted positives out of the total positives.
\begin{center}
    \( Precision (\%) = \frac{TP}{ TP + FP} \times 100 \) 
    \begin{spacing}{1.125}
    \end{spacing}
    \( Recall (\%)= \frac{TP}{ TP + FN} \times 100 \)
\end{center}
F1-score is the harmonic mean of precision and recall, which takes both false positives and false negatives into account. With the help of the confusion matrix, all important measures can be obtained and provide us with a better insight into the predictions made by MesoInception-4 model and the transformation in the input-output pairs of the model. 
\begin{center}
    \( F1-score(100\%) = \frac{2 \times Precision \times Recall } { Precision + Recall} \times 100\)
\end{center}

\newcommand{\boldm}[1] {\mathversion{bold}#1\mathversion{normal}}

\subsection{Setup}

\begin{table}[b]
    \centering
    \caption{Accuracy of MesoInception-4 on imbalanced, partial balanced, and balanced gender image from FaceForensics dataset. \textbf{ACC} indicates the accuracy.}
    \resizebox{0.478\textwidth}{!}{%
    \begin{tabular}{llll}
    \hline
    \multicolumn{1}{l|}{Category} & \multicolumn{1}{l|}{Imbalanced} & \multicolumn{1}{l|}{Partial Balance} & Balance \\ \hline
    \multicolumn{1}{l|}{ACC on Male (\%)} & \multicolumn{1}{l|}{83.35} & \multicolumn{1}{l|}{83.02} & 83.22 \\
    \multicolumn{1}{l|}{ACC on Female (\%)} & \multicolumn{1}{l|}{87.44} & \multicolumn{1}{l|}{87.08} & 87.22 \\
    \multicolumn{1}{l|}{Bias Factor} & \multicolumn{1}{l|}{4.09} & \multicolumn{1}{l|}{4.06} & 4.00 \\ \hline
    \end{tabular}%
    }
    \label{tab:preperation_testing}
\end{table}

Metamorphic relation is the defined transformation of the input that has a measurable effect on the output. We therefore define the metamorphic relations for the purpose of evaluating whether they hold after the transformation, which is the test oracle of MT. According to the general functionality of deepfake detectors, we are able to identify a metamorphic relation prior to conducting the experiments, where the following metamorphic relationship is proposed to be the foundation of our study:

\hangindent=0.35cm \boldm $MR_{01}$ Introduction of makeup (feminine features) in the testing images should not change the decisions made by MesoInception-4 model.

\noindent \boldm $MR_{01}$ assumes that the introduction of makeup in the testing images should not alter or influence the overall performance of MesoInception-4, when the training samples are balanced. The deepfake images provided by FaceForensics dataset are generated by automated face manipulation algorithms including Deepfake and Face2Face \cite{rossler2019faceforensics}. Moreover, we argue that makeup is common in the real-world, both in male and female subjects and hence, real images with makeup applied should not be considered as deepfake \cite{rossler2019faceforensics}. It is also reasonable to assume the model is designed to learn a similar amount of deepfake features from both genders to ensure fairness of the model since makeup is more common in female images. This metamorphic relation was proposed to evaluate if deepfake detection models are bias towards either feminine or masculine features. Thus, we find that it is natural to assume that applying makeup on the focal objects (males and females) will have no significant impact on the prediction results (predict whether the test image is deepfake or not). 


We first train on the original dataset with imbalanced samples for different genders and then proved the assumption that imbalanced gender data will indeed cause unfair results from MesoInception-4 model, as shown in Table \ref{tab:preperation_testing}. However, when balanced training samples are fed into the model, the model still produces unfair decisions on both genders, illustrated in Table \ref{tab:preperation_testing}. To measure if the model is bias towards one gender over the other, we define the term \textit{bias factor} to measure the fairness of the model between female and male subjects. \textit{Bias factor} is an absolute value, which is calculated as the performance difference between female and male measured by their accuracy. For example, if we obtain an accuracy of 86.36\% for the model to correctly detect male deepfake images and 90.1\% for female deepfake images, then this experiment resulted in a bias factor of 3.74\%. Based on the results, data bias is therefore not to be blamed for the model's unfair in performance. At this point, the accuracy shown in the first row of Table \ref{tab:Metrixs_Summary} closely resembles the accuracy reported in MesoInception-4 and hence, it can be confidently used as the ground truth for our study.

\begin{figure}[ht]
    \centering
    \begin{subfigure}[b]{0.115\textwidth}
        \caption*{Original sample}
        \includegraphics[width=\textwidth]{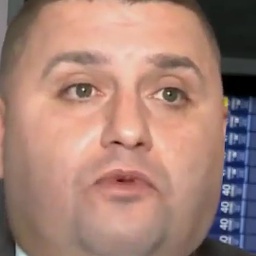}
    \end{subfigure}
    ~
    \begin{subfigure}[b]{0.115\textwidth}
        \caption*{$TC_{01}$}
        \includegraphics[width=\textwidth]{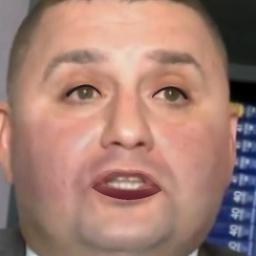}
    \end{subfigure}
    ~
    \begin{subfigure}[b]{0.115\textwidth}
        \caption*{$TC_{02}$}
        \includegraphics[width=\textwidth]{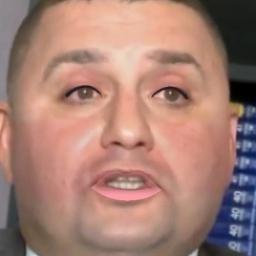}
    \end{subfigure}
    ~ 
    \begin{subfigure}[b]{0.115\textwidth}
        \caption*{$TC_{03}$}
        \includegraphics[width=\textwidth]{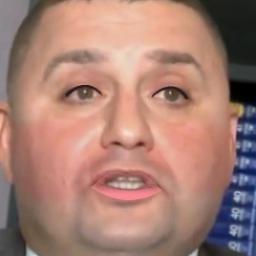}
    \end{subfigure}
    \\
    \begin{subfigure}[b]{0.115\textwidth}
        \caption*{$TC_{04}$}
        \includegraphics[width=\textwidth]{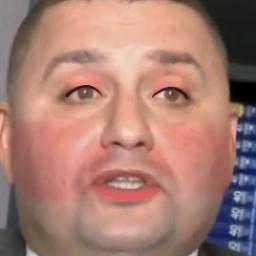}
    \end{subfigure}
    ~
    \begin{subfigure}[b]{0.115\textwidth}
        \caption*{$TC_{05}$}
        \includegraphics[width=\textwidth]{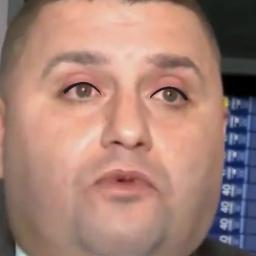}
    \end{subfigure}
    ~
    \begin{subfigure}[b]{0.115\textwidth}
        \caption*{$TC_{06}$}
        \includegraphics[width=\textwidth]{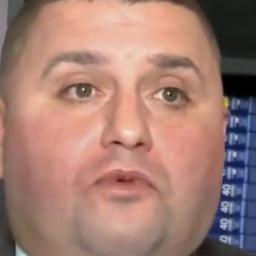}
    \end{subfigure}
    ~
    \begin{subfigure}[b]{0.115\textwidth}
        \caption*{$TC_{07}$}
        \includegraphics[width=\textwidth]{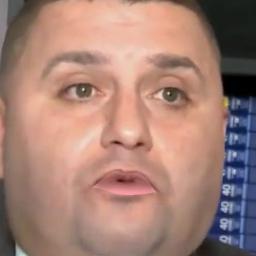}
    \end{subfigure}
    \caption{Samples of makeup images in different test cases with contrast to the original image.}
    \label{fig:TC_Summary}
\end{figure}

After establishing the fundamental metamorphic relationship, we began looking for makeup components to insert in the training images. We have chosen eyeliners, eyeshadows, blushes, and lipsticks since we want to investigate how the performance of MesoInception-4 changes when a specific gender's facial features are enhanced. The makeup is performed by identification of 68 facial landmarks through the usage of Dlib libraries followed by the interpolation of each makeup region boundaries. The last step of makeup involves filling in the suitable RGB values onto the interpolated regions to create the makeup components. These components need to be a suitable size and natural to the face of the focal objects in each image. Samples of the makeup images are presented in Figure \ref{fig:TC_Summary}, with the labels $TC_{01}$-$TC_{07}$ corresponding to the test cases derived from the various MRs which will be discussed in the following sections. Furthermore, Table \ref{tab:MR_Summary} includes a summary of the associated descriptions and test cases for all MRs that will be discussed in the following section.

\begin{figure}[hb]
    \centering
    \begin{subfigure}[b]{0.14\textwidth}
        \caption*{Step 1 - Allocate}
        \includegraphics[width=\textwidth]{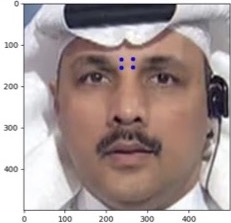}
    \end{subfigure}
    ~
    \begin{subfigure}[b]{0.14\textwidth}
        \caption*{Step 2 - Mark}
        \includegraphics[width=\textwidth]{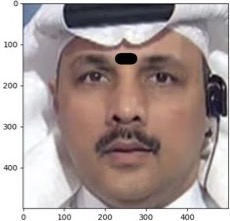}
    \end{subfigure}
    ~ 
    \begin{subfigure}[b]{0.14\textwidth}
        \caption*{Step 3 - Extract}
        \includegraphics[width=\textwidth]{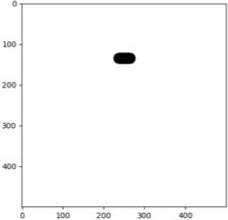}
    \end{subfigure}
    \caption{Steps to find the adaptive RGB value and intensity.}
    \label{fig:skin_detection}
\end{figure} 

\begin{table*}[t]
    \centering
    \caption{Summary of all MRs, their relationships, and derived TCs.}
    \resizebox{0.85\textwidth}{!}{%
    \begin{tabular}{l|llll}
    \hline
    MRs & \multicolumn{4}{l}{Information} \\ \hline
    \multirow{3}{*}{$MR_{01}$} & \multicolumn{1}{l|}{Description} & \multicolumn{3}{l}{Introduction of make-up in testing images should not change the decisions made by MesoInception-4.} \\ \cline{2-5} 
     & \multicolumn{1}{l|}{Causal Relation} & \multicolumn{3}{l}{-} \\ \cline{2-5} 
     & \multicolumn{1}{l|}{Test Case} & \multicolumn{1}{l|}{$TC_{01}$} & \multicolumn{1}{l|}{Modification} & Introduction of adaptive make-up to all the testing images. \\ \hline
    \multirow{5}{*}{$MR_{02}$} & \multicolumn{1}{l|}{Description} & \multicolumn{3}{l}{\begin{tabular}[c]{@{}l@{}}Introduction of different intensity of make-up, including light, medium \& heavy, should not change \\ the decision made by deepfake detector model.\end{tabular}} \\ \cline{2-5} 
     & \multicolumn{1}{l|}{Causal Relation} & \multicolumn{3}{l}{$MR_{01}$} \\ \cline{2-5} 
     & \multicolumn{1}{l|}{\multirow{3}{*}{Test Case}} & \multicolumn{1}{l|}{$TC_{02}$} & \multicolumn{1}{l|}{\multirow{3}{*}{Modification}} & Introduction of light intensity make-up to all the testing images. \\ \cline{3-3} \cline{5-5} 
     & \multicolumn{1}{l|}{} & \multicolumn{1}{l|}{$TC_{03}$} & \multicolumn{1}{l|}{} & Introduction of medium intensity of make-up to all the testing images. \\ \cline{3-3} \cline{5-5} 
     & \multicolumn{1}{l|}{} & \multicolumn{1}{l|}{$TC_{04}$} & \multicolumn{1}{l|}{} & Introduction of heavy intensity of make-up to all the testing images. \\ \hline
    \multirow{5}{*}{$MR_{03}$} & \multicolumn{1}{l|}{Description} & \multicolumn{3}{l}{\begin{tabular}[c]{@{}l@{}}Introduction of make-up with light intensity on the different facial regions should not change\\  the decision made by deepfake detector model.\end{tabular}} \\ \cline{2-5} 
     & \multicolumn{1}{l|}{Causal Relation} & \multicolumn{3}{l}{$MR_{01}$, $MR_{02}$} \\ \cline{2-5} 
     & \multicolumn{1}{l|}{\multirow{3}{*}{Test Case}} & \multicolumn{1}{l|}{$TC_{05}$} & \multicolumn{1}{l|}{\multirow{3}{*}{Modification}} & Introduction of light intensity eyeliner and eyeshadow to all the testing images. \\ \cline{3-3} \cline{5-5} 
     & \multicolumn{1}{l|}{} & \multicolumn{1}{l|}{$TC_{06}$} & \multicolumn{1}{l|}{} & Introduction of light intensity blush to all the testing images. \\ \cline{3-3} \cline{5-5} 
     & \multicolumn{1}{l|}{} & \multicolumn{1}{l|}{$TC_{07}$} & \multicolumn{1}{l|}{} & Introduction of light intensity lipstick to all the testing images. \\ \hline
    \end{tabular}%
    }
    \label{tab:MR_Summary}
\end{table*}

\subsection{Results and Discussion}
Prior to executing any test cases, we verified one important hypothesis that testing dataset with an imbalanced sample of both genders will not affect the bias factor of MesoInception-4. In all scenarios, we can see from Table \ref{tab:preperation_testing} that they have an approximate bias factor of 4\% towards female. 

For the sake of bias mitigation due to imbalance input data, we proceed to use the gender dataset that are fully balanced retrieved from FaceForensics, where the new results are summarised in the first and second rows of Table \ref{tab:Metrixs_Summary}. Numbers in bold are experiment results for the original sample used as benchmark.

To validate \boldm $MR_{01}$, we derive the first test case, \boldm $TC_{01}$, which investigates the ability of MesoInception-4 to handle the adaptive makeup that adds extra feminine features to each testing image. The term \textit{adaptive} implies making the makeup intensity adjusted by the original image’s color intensity and allow for manual modification of the RGB values. The workflow of finding the adaptive value can be seen in Figure \ref{fig:skin_detection}, where we first (1) locate at the position between eyebrows, then (2) generate a mask by drawing a rectangle using selected points, and finally (3) calculate the average RGB value and the mean intensity from the extracted area. 

\begin{figure}[b]
     \centering
     \includegraphics[width=0.478\textwidth]{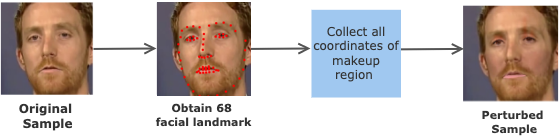}
     \caption{Detect facial landmarks and perturb the images.}
     \label{fig:makeupFlow}
\end{figure}

To further elaborate the components of adaptive makeup, we emphasis on facial regions which are eyes, cheeks, and lips as they can better exhibit one's personal features. Besides the adaptive approach of choosing makeup colors, each components are resized according to one's native face using a facial landmark shown in Figure \ref{fig:makeupFlow} and highlighted using red dots. 

Generally, all makeup applications involve a few common steps as stated in Figure \ref{fig:makeupFlow}. Several image statistics such as image size and height are first computed and the image is passed on to the Dlib facial landmarks detector to determine the coordinates of the 68 facial landmarks. Based on the information collected, the algorithm then retrieves only the required facial landmarks coordinates based on the face region indices. The coordinates are discrete and hence interpolation function is employed to connect them into a continuous boundary line of the desired makeup application region. Upon getting the outer line of the makeup region, all the interior coordinates are computed and stored together. The last two steps involve applying makeup with the given RGB values onto the makeup region, computed, and smoothened using Gaussian Filter blurring and image blending.

As stated in \boldm $MR_{01}$, we expect that the resulting adaptive makeup images would not affect the decision made by the MesoInception-4 model. We expect to see some tolerance of MesoInception-4 model with regard to the makeup components, as it is very common for people to wear makeup in the real-world. 

\begin{figure}[b]
    \centering
    \begin{subfigure}[b]{0.478\textwidth}
        \includegraphics[width=\textwidth]{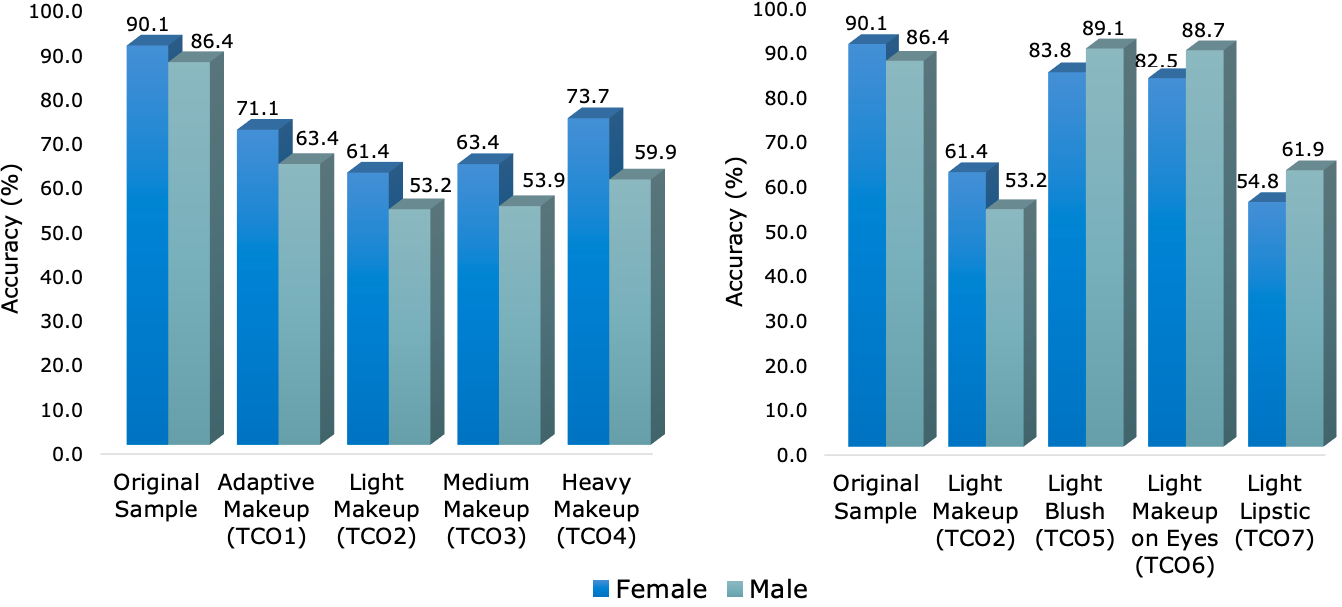}
    \end{subfigure}
    \caption{Accuracy for all test cases. The experiment results obtained from the original samples are shown in the first two columns (benchmark for both male and female images).}
    \label{fig:acc_barchart}
\end{figure}

However, the results did not match our expectations. Based on our experiment results, we found that the adaptive makeup does not maintain the model's performance and degrades the performance. We can observe from Figure \ref{fig:acc_barchart} that the accuracy for detecting male (63.36 \%) and female (71.1 \%) deepfake images both dropped by around 4\% compared to the benchmark. Theoretically, the accuracy for both gender should drop to a similar degree given that they are applying the same makeup function. Interestingly, we can see a significant drop in performance from Table \ref{tab:Metrixs_Summary}, where the recall value of detecting male deepfake images drops from 90.16\% to 28.3\%. Likewise, the recall value of female subjects declines from 92.02\% to 44.22\%. As opposed to recall, the precision of identifying deepfake images for both males and females shows an growth pattern. This is because fewer images are detected as deepfake, which further results an increase in the percentage of predicted deepfake samples out of the total deepfake samples. This transformation in the values of precision indicates that the adaptive makeup managed to confuse the model to consider a huge amount of deepfake images as real. Furthermore, the F1-score also corresponded to our previous judgement and showed a dramatic decrease in both genders. The decrease values with the introduction of adaptive makeup further revealed the gender fairness issue in MesoInception-4 model and the unstable performance of the model. 

The poor result from \boldm $MR_{01}$ means that the model may have misidentified the perturbed deepfake images as real. This result leads us to define a new metamorphic relation, \boldm $MR_{02}$, in order to reveal the reasons behind the decrease in MesoInception-4 performance (as shown in Table \ref{tab:MR_Summary}). We are curious about the influence of makeup with different intensity values and examine whether MesoInception-4 is more robust towards lighter makeup (which implies less feminine features), compared to heavier makeup. \boldm $MR_{02}$ was introduced to help evaluate if the makeup with an intensity value of different degree affect the model in a same manner. In the real-world, a person tends to wear different makeup for different occasions and hence, it is reasonable to assume that the intensity of makeup is not a factor cause the degradation of MesoInception-4.  

\begin{table}[b]
    \centering
    \caption{Summary of evaluation metrics for all test cases.}
    \resizebox{0.478\textwidth}{!}{%
    \begin{tabular}{l|l|l|l|l}
    \hline
    Test Case & {\color[HTML]{333333} Accuracy (\%)} & \multicolumn{1}{c|}{{\color[HTML]{333333} Recall (\%)}} & \multicolumn{1}{c|}{{\color[HTML]{333333} Precision (\%)}} & \multicolumn{1}{c}{{\color[HTML]{333333} F1-score (\%)}} \\ \hline
     & \cellcolor[HTML]{C0C0C0}{\color[HTML]{000000} \textbf{86.36}} & \cellcolor[HTML]{C0C0C0}{\color[HTML]{000000} \textbf{90.16}} & \cellcolor[HTML]{C0C0C0}{\color[HTML]{000000} \textbf{83.71}} & \cellcolor[HTML]{C0C0C0}{\color[HTML]{000000} \textbf{86.82}} \\
    \multirow{-2}{*}{Original Sample (Benchmark)} & \textbf{90.1} & \textbf{92.2} & \textbf{88.59} & \textbf{90.36} \\ \cline{1-1}
     & \cellcolor[HTML]{C0C0C0}{\color[HTML]{333333} 63.36} & \cellcolor[HTML]{C0C0C0}{\color[HTML]{333333} 28.3} & \cellcolor[HTML]{C0C0C0}{\color[HTML]{333333} 93.62} & \cellcolor[HTML]{C0C0C0}{\color[HTML]{333333} 43.46} \\
    \multirow{-2}{*}{$TC_{01}$ (Adaptive make-up)} & 71.1 & 44.22 & 95.61 & 60.47 \\ \cline{1-1}
     & \cellcolor[HTML]{C0C0C0}{\color[HTML]{333333} 53.16} & \cellcolor[HTML]{C0C0C0}{\color[HTML]{333333} 6.5} & \cellcolor[HTML]{C0C0C0}{\color[HTML]{333333} 94.11} & \cellcolor[HTML]{C0C0C0}{\color[HTML]{333333} 12.16} \\
    \multirow{-2}{*}{$TC_{02}$ (Light make-up)} & 61.43 & 24.34 & 97.96 & 38.99 \\ \cline{1-1}
     & \cellcolor[HTML]{C0C0C0}53.87 & \cellcolor[HTML]{C0C0C0}7.81 & \cellcolor[HTML]{C0C0C0}96.25 & \cellcolor[HTML]{C0C0C0}14.45 \\
    \multirow{-2}{*}{$TC_{03}$ (Medium make-up)} & 63.35 & 28.1 & 98.58 & 43.73 \\ \cline{1-1}
     & \cellcolor[HTML]{C0C0C0}{\color[HTML]{333333} 59.86} & \cellcolor[HTML]{C0C0C0}{\color[HTML]{333333} 21.51} & \cellcolor[HTML]{C0C0C0}{\color[HTML]{333333} 92.11} & \cellcolor[HTML]{C0C0C0}{\color[HTML]{333333} 34.88} \\
    \multirow{-2}{*}{$TC_{04}$  (Heavy make-up)} & 73.7 & 50.71 & 95.6 & 66.27 \\ \cline{1-1}
     & \cellcolor[HTML]{C0C0C0}83.78 & \cellcolor[HTML]{C0C0C0}86.21 & \cellcolor[HTML]{C0C0C0}82.21 & \cellcolor[HTML]{C0C0C0}84.16 \\
    \multirow{-2}{*}{$TC_{05}$ (Light blush)} & 89.11 & 90.67 & 87.94 & 89.28 \\ \cline{1-1}
     & \cellcolor[HTML]{C0C0C0}82.48 & \cellcolor[HTML]{C0C0C0}85.18 & \cellcolor[HTML]{C0C0C0}80.83 & \cellcolor[HTML]{C0C0C0}82.95 \\
    \multirow{-2}{*}{$TC_{06}$ (Light make-up on eyes)} & 88.7 & 89.84 & 87.84 & 88.83 \\ \cline{1-1}
     & \cellcolor[HTML]{C0C0C0}{\color[HTML]{333333} 54.76} & \cellcolor[HTML]{C0C0C0}{\color[HTML]{333333} 10.15} & \cellcolor[HTML]{C0C0C0}{\color[HTML]{333333} 94.23} & \cellcolor[HTML]{C0C0C0}{\color[HTML]{333333} 18.33} \\
    \multirow{-2}{*}{$TC_{07}$ (Light lipstick)} & 61.86 & 33.36 & 97.57 & 49.72 \\ \hline
    \end{tabular}%
    }
    \\
    \begin{subfigure}[b]{0.15\textwidth}
        \includegraphics[width=\textwidth]{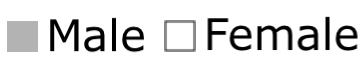}
    \end{subfigure}
    \label{tab:Metrixs_Summary}
\end{table}

\hangindent=0.35cm \boldm $MR_{02}$ state that the introduction of different intensities of makeup, including light, medium, and heavy should not change the decision made by MesoInception-4.

\noindent Visually, the dataset images contain human subjects with a variety of skin tones, where we categorise them into 3 main categories including light, medium and heavy. Based on the categorised skin tone range, we defined a fixed intensity value for different makeup levels in each skin tone category to improve the makeup application outcomes. 

The second test case, \boldm $TC_{02}$, was built based on the \boldm $MR_{02}$, where we hope to see similar effect as observed in \boldm $TC_{01}$. Figure \ref{fig:acc_barchart} shows the results of \boldm $TC_{02}$. While the result was inconsistent with $MR_{02}$, we observed that the lighter makeup from $TC_{02}$ cause the model to perform much worse than $TC_{01}$. The accuracy dropped by roughly 10\% in $TC_{02}$ for both genders, while there is less impact on the bias factor with an increase of 0.53\% towards female. This finding is unexpected because the addition of light makeup nearly break the model, resulting the model to randomly guess a decision, especially for male images. Based on the results from our two test cases, where the result of $TC_{01}$ was moderately better than $TC_{02}$, this implies that the intensity value of makeup does matter and will impact the model differently. 

Following this, the third and fourth test case \boldm $TC_{03}$ \& \boldm $TC_{04}$ was proposed to evaluate if a higher intensity value leads to a worse performance or makes the model to produce less fair outcomes.

The result of \boldm $TC_{03}$ shown in Figure \ref{fig:acc_barchart} were however, found to be unfaithful to $MR_{02}$ as well, similar to its predecessor, $TC_{02}$. Despite this, we were able to interpret transformation pattern existed in Figure \ref{fig:acc_barchart}, where the intensity value is inversely proportional to the accuracy of the model but proportional to the the bias factor. When we change the test case from light makeup to medium makeup, the bias factor increase from 4.53\% to 5.79\% with a decrease of roughly 10\% in terms of accuracy compared to that value on our original input data. At this point, we believe the intensity value has great effect on the model as the two test cases, $TC_{02}$ and $TC_{03}$, provide us a relatively reliable result (see Figure \ref{fig:acc_barchart} and Table \ref{tab:Metrixs_Summary}). This effect will be further investigated in the follow-up test case, $TC_{04}$. 

\begin{figure}[t]
    \centering
    \begin{subfigure}[b]{0.478\textwidth}
        \includegraphics[width=\textwidth]{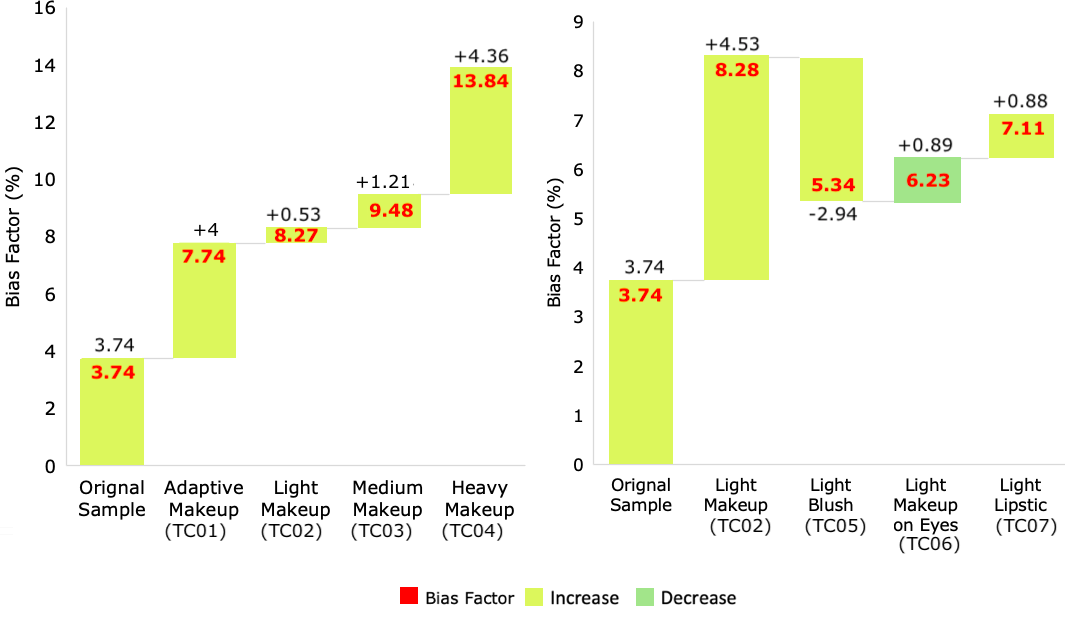}
    \end{subfigure}
    \caption{Bias factor for all test cases. The values are obtained by taking the absolute difference of the accuracy to detect male and female images. The results for the original samples which act as the benchmark are shown in the first column of the two graphs above.}
    \label{fig:bias_waterfall}
\end{figure}

Having now established that makeup intensities are very likely to be the main contributing factor, we intended to discover a clearer pattern of transformation from the test results. Visually, the difference between light and medium makeup is not very obvious. With the interest to pursue the irregularity of $MR_{02}$ further, we enlarged the difference between medium and heavy makeup with $TC_{04}$ (Table \ref{tab:MR_Summary}). By referring to Figure \ref{fig:acc_barchart} and Table \ref{tab:Metrixs_Summary}, $TC_{04}$ resulted in higher accuracy than both $TC_{01}$ and $TC_{02}$, possibly because heavy makeup is more obvious for the model to comprehend and learn compared to light and medium makeup. The trend of accuracy that illustrated in Figure \ref{fig:acc_barchart} confirmed our assumption. The accuracy on both male and female samples underwent a continuous decrease from adaptive makeup to light makeup, then increased faintly in $TC_{03}$. Finally, in $TC_{04}$, the performance of the model grew again with the accuracy of 59.86\% for male and 73.7\% for female. This transformation of accuracy also appeared in the remaining evaluation metrics for the both genders in $TC_{04}$. At this stage of our experiments, we can claim that all the test cases proposed for $MR_{02}$ are in contradiction with the claim of $MR_{02}$, where we assumed that the increase in makeup intensity will not cause the deterioration of the model's performance and cause the model to perform less fairly due to the introduction of more profound feminine features. 

Moving on from makeup intensity, the impact of makeup can also be attributed to facial regions. The third metamorphic relation, $MR_{03}$, is therefore proposed to further explore this impact. 

\hangindent=0.35cm \boldm $MR_{03}$ Introduction of makeup with the same intensity on different facial regions should not change the decision made by MesoInception-4. 

\noindent Its corresponding test cases $TC_{05}$, $TC_{06}$ and $TC_{07}$ (refer to Table \ref{tab:MR_Summary}) are derived to observe how would these test cases impact the MesoInception-4 model with respect to the three major facial regions including eyes ($TC_{05}$), cheeks ($TC_{06}$), and lips ($TC_{07}$) based on the testing results of $MR_{01}$ and $MR_{02}$ where light intensity makeup lead the model to perform at its worst in accordance to both $MR_{01}$ and $MR_{02}$. In order to ensure the reliability and consistency of our results, all the test cases for $MR_{03}$ were implemented using the same intensity level, that is light intensity. We decided to choose the light level of intensity because it posted the most severe influence on the model's performance (see Figure \ref{fig:acc_barchart}). We were hence interested to identify an easily observable effect that reflects on the performance of adding makeup on selected facial regions. 

At this stage of our experiments, we expect to verify that $MR_{03}$ is applicable to the test case no matter the position of makeup. However, for both male and female images, the accuracy of $TC_{05}$ almost similar to our benchmark, but the results are much better than $TC_{02}$. We now have the reason to believe that there is a certain relationship between the model's performance and the position of the makeup, which contradict with the statement of $MR_{03}$. Consistently, there is only a slight decrease in the performance of the model for $TC_{06}$ compare to that of $TC_{05}$ with an accuracy of 82.48 \% for male images and 88.7 \% for female images. This finding is interesting because $TC_{05}$ and $TC_{06}$ achieved relatively similar results compared to the benchmark. We now move on to our last test case $TC_{07}$, and the results strongly go against $MR_{03}$. Light makeup on the lips confuses the model further when compared to cheeks and eyes, and brings the testing result of $TC_{07}$ nearly back to that of $TC_{02}$ (see Figure \ref{fig:acc_barchart}). This observation indicate a salient region on or near the lips that will bring the greatest impact on the MesoInception-4 model and the fairness of it among all the test cases constructed for $MR_{03}$ (see Figure \ref{fig:bias_waterfall}).

Finally, we review the results of Figure \ref{fig:acc_barchart} \& \ref{fig:bias_waterfall} and examine them in a qualitative manner. We believe that some anomalies appeared in all test cases in a consistent manner such that: 
\begin{itemize}
    \item The MesoInception-4 model generally produces higher deepfake detection accuracy for female images, showing that it is more favourable towards female subjects.
    \item Makeup is not tolerated by the MesoInception-4 model and would further detract from the fairness of the model. 
    \item If left unattended, malicious users can exploit the vulnerability of lipstick makeup to fool the MesoInception-4 model, rendering it not able to accurately detect deepfake images.  
\end{itemize}

\section{Conclusion and Future Work}
MT empowers its users to address the software testing challenges of machine learning and DL-based systems, where these kinds of systems normally operate on fairly large input spaces with probabilistic outcomes from mostly black-box components. 
In this paper, we verified that MT is fairly effective in determining the existence of gender bias in the state-of-the-art deepfake detection model, MesoInception-4. Metamorphic relations are derived based on the assumption that makeup on both male and female subjects are expected behaviour, and it should not interfere with the models' interpretation. Any deviation of the models' output due to the makeup could signify a bias in gender due to more feminine or masculine features in the image. Our test results prove that the MesoInception-4 model is generally unfair towards both genders, with more preferences towards female. At the same time, we reveal how makeup perturbation can be used as an adversarial attack technique to degrade and fool the model. We have also exposed the impact of different makeup levels on the model and identified the salient region which causes the highest performance degradation. Not only does the series of MT conducted shows the gender bias property of MesoInception-4 model, we have also revealed one of its vulnerabilities - that we can fool the model easily with makeup perturbations. In conclusion, we hope that these findings could create awareness among the AI communities about having an improved algorithm to mitigate gender bias. State-of-the-art deepfake detection models should also have the ability to tolerate the makeup applied to the test subject, which is a norm in the society nowadays regardless of genders. We also believe that MT can be further utilised in the continuous effort to address the test oracle problem of deepfake detectors in other aspects.

The current work focuses solely on images with compression rate of 23 and deepfake images generated using the Face2face technique. Future studies may consider experimenting with images using varying compression rates and deepfake techniques, which could reveal additional insights of the proposed MT approach. Besides, it will be interesting to see if fairness issues exist in other state-of-the-art deepfake detection models as well, apart from MesoInception-4.

\begin{acks}
This work was supported by the Advanced Engineering Platform's Cluster Funding (AEP-2021-Cluster-04), Monash University, Malaysia.
\end{acks}

\bibliographystyle{ACM-Reference-Format}
\bibliography{reference}


\begin{thebibliography}{15}


\ifx \showCODEN    \undefined \def \showCODEN     #1{\unskip}     \fi
\ifx \showDOI      \undefined \def \showDOI       #1{#1}\fi
\ifx \showISBNx    \undefined \def \showISBNx     #1{\unskip}     \fi
\ifx \showISBNxiii \undefined \def \showISBNxiii  #1{\unskip}     \fi
\ifx \showISSN     \undefined \def \showISSN      #1{\unskip}     \fi
\ifx \showLCCN     \undefined \def \showLCCN      #1{\unskip}     \fi
\ifx \shownote     \undefined \def \shownote      #1{#1}          \fi
\ifx \showarticletitle \undefined \def \showarticletitle #1{#1}   \fi
\ifx \showURL      \undefined \def \showURL       {\relax}        \fi
\providecommand\bibfield[2]{#2}
\providecommand\bibinfo[2]{#2}
\providecommand\natexlab[1]{#1}
\providecommand\showeprint[2][]{arXiv:#2}

\bibitem[\protect\citeauthoryear{Afchar, Nozick, Yamagishi, and Echizen}{Afchar
  et~al\mbox{.}}{2018}]%
        {MesoNet}
\bibfield{author}{\bibinfo{person}{Darius Afchar}, \bibinfo{person}{Vincent
  Nozick}, \bibinfo{person}{Junichi Yamagishi}, {and} \bibinfo{person}{Isao
  Echizen}.} \bibinfo{year}{2018}\natexlab{}.
\newblock \showarticletitle{MesoNet: a Compact Facial Video Forgery Detection
  Network}.
\newblock \bibinfo{journal}{\emph{2018 IEEE International Workshop on
  Information Forensics and Security (WIFS)}} (\bibinfo{date}{Dec}
  \bibinfo{year}{2018}).
\newblock


\bibitem[\protect\citeauthoryear{Chollet}{Chollet}{2017}]%
        {XceptionNet}
\bibfield{author}{\bibinfo{person}{Fran{\c{c}}ois Chollet}.}
  \bibinfo{year}{2017}\natexlab{}.
\newblock \showarticletitle{Xception: Deep learning with depthwise separable
  convolutions}. In \bibinfo{booktitle}{\emph{Proceedings of the IEEE
  Conference on Computer Vision and Pattern Recognition}}.
  \bibinfo{pages}{1251--1258}.
\newblock


\bibitem[\protect\citeauthoryear{Ding, Raziei, Larson, Olinick, Krueger, and
  Hahsler}{Ding et~al\mbox{.}}{2020}]%
        {ding_raziei_larson_olinick_krueger_hahsler_2020}
\bibfield{author}{\bibinfo{person}{Xinyi Ding}, \bibinfo{person}{Zohreh
  Raziei}, \bibinfo{person}{Eric~C. Larson}, \bibinfo{person}{Eli~V. Olinick},
  \bibinfo{person}{Paul Krueger}, {and} \bibinfo{person}{Michael Hahsler}.}
  \bibinfo{year}{2020}\natexlab{}.
\newblock \bibinfo{title}{Swapped face detection using Deep Learning and
  Subjective Assessment - EURASIP Journal on Information Security}.
\newblock
\newblock


\bibitem[\protect\citeauthoryear{Li, Bao, Zhang, Yang, Chen, Wen, and Guo}{Li
  et~al\mbox{.}}{2020}]%
        {li2020face}
\bibfield{author}{\bibinfo{person}{Lingzhi Li}, \bibinfo{person}{Jianmin Bao},
  \bibinfo{person}{Ting Zhang}, \bibinfo{person}{Hao Yang},
  \bibinfo{person}{Dong Chen}, \bibinfo{person}{Fang Wen}, {and}
  \bibinfo{person}{Baining Guo}.} \bibinfo{year}{2020}\natexlab{}.
\newblock \showarticletitle{Face X-Ray for More General Face Forgery
  Detection}. In \bibinfo{booktitle}{\emph{2020 IEEE/CVF Conference on Computer
  Vision and Pattern Recognition (CVPR)}}.
\newblock


\bibitem[\protect\citeauthoryear{Mehrabi, Morstatter, Saxena, Lerman, and
  Galstyan}{Mehrabi et~al\mbox{.}}{2021}]%
        {mehrabi2021survey}
\bibfield{author}{\bibinfo{person}{Ninareh Mehrabi}, \bibinfo{person}{Fred
  Morstatter}, \bibinfo{person}{Nripsuta Saxena}, \bibinfo{person}{Kristina
  Lerman}, {and} \bibinfo{person}{Aram Galstyan}.}
  \bibinfo{year}{2021}\natexlab{}.
\newblock \showarticletitle{A survey on bias and fairness in machine learning}.
\newblock \bibinfo{journal}{\emph{ACM Computing Surveys (CSUR)}}
  \bibinfo{volume}{54}, \bibinfo{number}{6} (\bibinfo{year}{2021}),
  \bibinfo{pages}{1--35}.
\newblock


\bibitem[\protect\citeauthoryear{Mekala, Porter, and Lindvall}{Mekala
  et~al\mbox{.}}{2020}]%
        {MTonMultiLevelFaceRecognitionModel}
\bibfield{author}{\bibinfo{person}{Rohan~Reddy Mekala}, \bibinfo{person}{Adam
  Porter}, {and} \bibinfo{person}{Mikael Lindvall}.}
  \bibinfo{year}{2020}\natexlab{}.
\newblock \showarticletitle{Metamorphic Filtering of Black-Box Adversarial
  Attacks on Multi-Network Face Recognition Models}. In
  \bibinfo{booktitle}{\emph{Proceedings of the IEEE/ACM 42nd International
  Conference on Software Engineering Workshops}}. \bibinfo{pages}{410–417}.
\newblock
\showISBNx{9781450379632}


\bibitem[\protect\citeauthoryear{Neekhara, Dolhansky, Bitton, and
  Ferrer}{Neekhara et~al\mbox{.}}{2021}]%
        {AdvAttackonDeepfakeDetector}
\bibfield{author}{\bibinfo{person}{Paarth Neekhara}, \bibinfo{person}{Brian
  Dolhansky}, \bibinfo{person}{Joanna Bitton}, {and}
  \bibinfo{person}{Cristian~Canton Ferrer}.} \bibinfo{year}{2021}\natexlab{}.
\newblock \showarticletitle{Adversarial threats to deepfake detection: A
  practical perspective}. In \bibinfo{booktitle}{\emph{Proceedings of the
  IEEE/CVF Conference on Computer Vision and Pattern Recognition}}.
  \bibinfo{pages}{923--932}.
\newblock


\bibitem[\protect\citeauthoryear{Ntoutsi, Fafalios, Gadiraju, Iosifidis, Nejdl,
  Vidal, Ruggieri, Turini, Papadopoulos, Krasanakis, Kompatsiaris,
  Kinder-Kurlanda, Wagner, Karimi, Fernandez, Alani, Berendt, Kruegel, Heinze,
  Broelemann, Kasneci, Tiropanis, and Staab}{Ntoutsi et~al\mbox{.}}{2020}]%
        {https://doi.org/10.1002/widm.1356}
\bibfield{author}{\bibinfo{person}{Eirini Ntoutsi}, \bibinfo{person}{Pavlos
  Fafalios}, \bibinfo{person}{Ujwal Gadiraju}, \bibinfo{person}{Vasileios
  Iosifidis}, \bibinfo{person}{Wolfgang Nejdl}, \bibinfo{person}{Maria-Esther
  Vidal}, \bibinfo{person}{Salvatore Ruggieri}, \bibinfo{person}{Franco
  Turini}, \bibinfo{person}{Symeon Papadopoulos}, \bibinfo{person}{Emmanouil
  Krasanakis}, \bibinfo{person}{Ioannis Kompatsiaris},
  \bibinfo{person}{Katharina Kinder-Kurlanda}, \bibinfo{person}{Claudia
  Wagner}, \bibinfo{person}{Fariba Karimi}, \bibinfo{person}{Miriam Fernandez},
  \bibinfo{person}{Harith Alani}, \bibinfo{person}{Bettina Berendt},
  \bibinfo{person}{Tina Kruegel}, \bibinfo{person}{Christian Heinze},
  \bibinfo{person}{Klaus Broelemann}, \bibinfo{person}{Gjergji Kasneci},
  \bibinfo{person}{Thanassis Tiropanis}, {and} \bibinfo{person}{Steffen
  Staab}.} \bibinfo{year}{2020}\natexlab{}.
\newblock \showarticletitle{Bias in data-driven artificial intelligence
  systems—An introductory survey}.
\newblock \bibinfo{journal}{\emph{WIREs Data Mining and Knowledge Discovery}}
  \bibinfo{volume}{10}, \bibinfo{number}{3} (\bibinfo{year}{2020}),
  \bibinfo{pages}{e1356}.
\newblock


\bibitem[\protect\citeauthoryear{P and Sk}{P and Sk}{2021}]%
        {9544522}
\bibfield{author}{\bibinfo{person}{Swathi P} {and} \bibinfo{person}{Saritha
  Sk}.} \bibinfo{year}{2021}\natexlab{}.
\newblock \showarticletitle{DeepFake Creation and Detection:A Survey}. In
  \bibinfo{booktitle}{\emph{2021 Third International Conference on Inventive
  Research in Computing Applications (ICIRCA)}}. \bibinfo{pages}{584--588}.
\newblock


\bibitem[\protect\citeauthoryear{Park, Waseem, Teo, Hwei~Low, Lim, and
  Yong~Chong}{Park et~al\mbox{.}}{2021}]%
        {StackedGenerativeAdversarialNetworksEvaluation}
\bibfield{author}{\bibinfo{person}{Hyejin Park}, \bibinfo{person}{Taaha
  Waseem}, \bibinfo{person}{Wen~Qi Teo}, \bibinfo{person}{Ying Hwei~Low},
  \bibinfo{person}{Mei~Kuan Lim}, {and} \bibinfo{person}{Chun Yong~Chong}.}
  \bibinfo{year}{2021}\natexlab{}.
\newblock \showarticletitle{Robustness Evaluation of Stacked Generative
  Adversarial Networks using Metamorphic Testing}.
\newblock \bibinfo{journal}{\emph{2021 IEEE/ACM 6th International Workshop on
  Metamorphic Testing (MET)}} (\bibinfo{date}{Jun} \bibinfo{year}{2021}).
\newblock


\bibitem[\protect\citeauthoryear{Rossler, Cozzolino, Verdoliva, Riess, Thies,
  and Niessner}{Rossler et~al\mbox{.}}{2019}]%
        {rossler2019faceforensics}
\bibfield{author}{\bibinfo{person}{Andreas Rossler}, \bibinfo{person}{Davide
  Cozzolino}, \bibinfo{person}{Luisa Verdoliva}, \bibinfo{person}{Christian
  Riess}, \bibinfo{person}{Justus Thies}, {and} \bibinfo{person}{Matthias
  Niessner}.} \bibinfo{year}{2019}\natexlab{}.
\newblock \showarticletitle{FaceForensics++: Learning to Detect Manipulated
  Facial Images}. In \bibinfo{booktitle}{\emph{Proceedings of the IEEE/CVF
  International Conference on Computer Vision (ICCV)}}.
\newblock


\bibitem[\protect\citeauthoryear{Szegedy, Liu, Jia, Sermanet, Reed, Anguelov,
  Erhan, Vanhoucke, and Rabinovich}{Szegedy et~al\mbox{.}}{2015}]%
        {szegedy2014going}
\bibfield{author}{\bibinfo{person}{Christian Szegedy}, \bibinfo{person}{Wei
  Liu}, \bibinfo{person}{Yangqing Jia}, \bibinfo{person}{Pierre Sermanet},
  \bibinfo{person}{Scott Reed}, \bibinfo{person}{Dragomir Anguelov},
  \bibinfo{person}{Dumitru Erhan}, \bibinfo{person}{Vincent Vanhoucke}, {and}
  \bibinfo{person}{Andrew Rabinovich}.} \bibinfo{year}{2015}\natexlab{}.
\newblock \showarticletitle{Going Deeper With Convolutions}. In
  \bibinfo{booktitle}{\emph{Proceedings of the IEEE Conference on Computer
  Vision and Pattern Recognition (CVPR)}}.
\newblock


\bibitem[\protect\citeauthoryear{Tan and Le}{Tan and Le}{2019}]%
        {EfficientNet}
\bibfield{author}{\bibinfo{person}{Mingxing Tan} {and} \bibinfo{person}{Quoc
  Le}.} \bibinfo{year}{2019}\natexlab{}.
\newblock \showarticletitle{Efficientnet: Rethinking model scaling for
  convolutional neural networks}. In \bibinfo{booktitle}{\emph{International
  Conference on Machine Learning}}. PMLR, \bibinfo{pages}{6105--6114}.
\newblock


\bibitem[\protect\citeauthoryear{Trinh and Liu}{Trinh and Liu}{2021}]%
        {trinh2021examination}
\bibfield{author}{\bibinfo{person}{Loc Trinh} {and} \bibinfo{person}{Yan Liu}.}
  \bibinfo{year}{2021}\natexlab{}.
\newblock \showarticletitle{An Examination of Fairness of AI Models for
  Deepfake Detection}. In \bibinfo{booktitle}{\emph{Proceedings of the
  Thirtieth International Joint Conference on Artificial Intelligence,
  {IJCAI-21}}}. \bibinfo{publisher}{International Joint Conferences on
  Artificial Intelligence Organization}, \bibinfo{pages}{567--574}.
\newblock


\bibitem[\protect\citeauthoryear{Wang and Su}{Wang and Su}{2020}]%
        {ODMT}
\bibfield{author}{\bibinfo{person}{Shuai Wang} {and} \bibinfo{person}{Zhendong
  Su}.} \bibinfo{year}{2020}\natexlab{}.
\newblock \showarticletitle{Metamorphic object insertion for testing object
  detection systems}. In \bibinfo{booktitle}{\emph{35th IEEE/ACM International
  Conference on Automated Software Engineering (ASE)}}. IEEE,
  \bibinfo{pages}{1053--1065}.
\newblock


\end{thebibliography}

\end{document}